\documentclass{article}

\usepackage[T1]{fontenc}
\usepackage[utf8]{inputenc}
\usepackage{lmodern}
\usepackage[margin=1in]{geometry}
\usepackage{amsmath,amssymb,bm}
\usepackage{booktabs,multirow,array,longtable,tabularx}
\usepackage{graphicx}
\usepackage{xcolor}
\usepackage{hyperref}
\usepackage{tikz}
\usepackage{pgfplots}
\usepackage{subcaption}
\usepackage{float}
\usepackage{enumitem}
\usepackage{authblk}
\usepackage{fancyhdr}
\usepackage{titlesec}
\pgfplotsset{compat=1.18}
\usetikzlibrary{arrows.meta,positioning,fit,shapes.geometric,calc,backgrounds}
\hypersetup{colorlinks=true,linkcolor=blue!50!black,citecolor=blue!50!black,urlcolor=blue!50!black}
\setlist[itemize]{leftmargin=1.5em}
\setlist[enumerate]{leftmargin=1.7em}
\newcolumntype{Y}{>{\raggedright\arraybackslash}X}

\titleformat{\section}{\Large\bfseries}{\thesection}{0.6em}{}
\titleformat{\subsection}{\large\bfseries}{\thesubsection}{0.6em}{}

\makeatletter
\renewcommand{\maketitle}{%
  \begingroup
    \begin{flushleft}
      {\LARGE\bfseries \@title\par}
      \vspace{1.2em}
      {\large \@author\par}
      \vspace{0.6em}
      {\small \@date\par}
    \end{flushleft}
  \endgroup
  \vspace{0.6em}
  \noindent\rule{\linewidth}{0.4pt}\par\vspace{0.8em}
}
\makeatother

\title{ChronoMedicalWorld: A Medical World Model for Learning Patient Trajectories from Longitudinal Care Data}

\author[1]{Jiangyuan Wang}
\author[1]{Xuyong Chen}
\author[1]{Junwei He}
\author[1]{Xu Xu}
\author[1]{Shasha Xie}
\author[1]{Fuman Han}
\affil[1]{Beijing KidneyTec Medical Technology Co., Ltd., Beijing, China}

\date{May 2026}

\newcommand{\smooth}{\operatorname{SmoothL1}}
\newcommand{\sg}{\operatorname{sg}}
\newcommand{\cmwm}{CMWM}

\begin{document}
\sloppy
\setlength{\emergencystretch}{3em}
\maketitle

\begin{abstract}
Long-horizon clinical simulation---predicting how a patient's physiology evolves over years under specified interventions---is central to chronic-disease care, yet existing electronic health record (EHR) models are predominantly discriminative, and general-purpose large language models drift under repeated interventions. We propose the \textbf{ChronoMedicalWorld Model (CMWM)}, an action-conditioned latent world-model framework for learning patient trajectories from longitudinal care data. CMWM couples a joint-embedding state encoder with a wide action encoder that admits both structured intervention indicators and free-text communication embeddings, and trains a recurrent latent transition module under a six-term objective: next-observation supervision, next-latent prediction, SIGReg latent regularisation, and three physiology-aware shape priors (slope, continuity, large-jump penalty). A closed-loop rollout-prefix protocol matches training to deployment, so the model is optimised against the same multi-step error it exhibits at inference. As a concrete case study, we instantiate CMWM for annual estimated glomerular filtration rate (eGFR) trajectory forecasting in chronic kidney disease (CKD). On a 2{,}232-patient nephrology cohort, the CKD instantiation achieves a dynamic-50\% history rollout test mean absolute error (MAE) of 7.384 and root-mean-square error (RMSE) of 10.256, against 7.964 and 11.069 for a tuned GPT-5.5 structured-prompting baseline ($-7.28\%$ MAE, $-7.35\%$ RMSE), with the gain dominated by the dialogue portion of patient--health-coach communication. The framework is not CKD-specific: its architecture, loss design, and training protocol apply to any chronic condition that can be cast as periodic clinical state interleaved with structured and conversational interventions.
\end{abstract}

\section{Introduction}
\label{sec:intro}

Care for chronic disease unfolds over years and is shaped by a continuous interplay between physiology and intervention: medications are titrated, lifestyle plans are negotiated, follow-up visits are scheduled, and conversational exchanges between patients and their care team---clinicians, nurses, or platform health coaches---communicate adherence, symptoms, and management decisions that are often only loosely reflected in structured laboratory tables. Modelling this process well would unlock individualised long-horizon trajectory forecasting, counterfactual intervention evaluation, and the broader vision of clinical digital twins. In practice, however, this capability is under-served. Classical risk equations \cite{tangri2011kfre} are calibrated for endpoint probabilities, not for multi-year state trajectories. Deep EHR representation learners such as RETAIN \cite{choi2016retain}, BEHRT \cite{li2020behrt}, and Med-BERT \cite{rasmy2021medbert}, along with disease-specific dynamic models \cite{yu2024tdlstm,ferro2024transformerckd,ma2025kfdeep}, are discriminative predictors of future events rather than closed-loop simulators of patient physiology. General-purpose large language models (LLMs), when prompted with rich clinical context, drift after a few intervention events---as quantified explicitly by recent patient-centric medical world-model work \cite{mu2026ehrworld}---because they lack a persistent latent state and were never trained against multi-step rollout error.

The world-model paradigm \cite{ha2018worldmodels,hafner2019planet,hafner2023dreamerv3,lecun2022pathway} offers an alternative: explicitly learn how observations evolve under actions, then unroll the learned dynamics in imagination to evaluate counterfactual intervention sequences. Joint-embedding predictive architectures (JEPAs) \cite{assran2023ijepa} place these dynamics in a compact latent space, and LeWorldModel (LeWM) \cite{maes2026leworldmodel} recently showed that a JEPA can be trained stably end-to-end with a next-embedding loss plus a sketched isotropic-Gaussian regulariser (SIGReg), removing many collapse-prevention heuristics. EHRWorld \cite{mu2026ehrworld} ported the world-model framing to medicine and demonstrated that disease-agnostic LLM-only simulators are not enough.

We identify three gaps in the current medical-world-model literature. \emph{First}, existing patient-centric medical world models are oriented toward broad clinical narrative simulation, and there is no published architectural recipe specialised to the periodic-observation regime in which structured interventions and free-text interventions co-occur, as is typical of chronic-disease follow-up. \emph{Second}, the action interface in medical-world-model work has so far treated interventions almost exclusively as structured events; the rich patient--health-coach text exchange that often carries the actual management signal has not been formalised as a first-class action input. \emph{Third}, the training-evaluation mismatch in which models are trained on next-step prediction but deployed in multi-step rollout is a known source of drift, yet it is not consistently addressed in clinical work.

This paper introduces the \textbf{ChronoMedicalWorld Model (\cmwm)}, a general action-conditioned latent world-model framework designed to fill these three gaps. \cmwm{} is not a CKD-specific predictor: it is a template that consumes (i) periodic clinical state observations of arbitrary structured dimensionality, (ii) action observations that may concatenate any structured intervention indicators with any free-text-derived semantic embeddings, and (iii) a target-time covariate vector; and produces both an explicit next-observation prediction and a next-latent prediction for closed-loop rollout. The framework is paired with a six-term training objective whose three physiology-aware shape priors (slope, continuity, jump) make it suitable for slowly evolving biomarkers, and with a rollout-prefix training protocol that closes the train--evaluation gap. To make the proposal concrete and to substantiate its value, we instantiate \cmwm{} for annual eGFR trajectory forecasting in CKD and compare it against a contemporary GPT-5.5 structured-prompting baseline.

Our contributions are:
\begin{itemize}
    \item \textbf{A medical-world-model framework.} We formalise long-horizon clinical trajectory forecasting under structured and conversational interventions as an action-conditioned latent dynamics problem and provide \cmwm, a concrete general-purpose recipe for it: state encoder, wide action encoder, recurrent latent transition module, and explicit next-observation and next-latent heads.
    \item \textbf{A first-class communication-action channel.} \cmwm{} treats patient--health-coach text exchange as an intervention signal rather than as auxiliary metadata, encoding each follow-up window's full transcript as a semantic vector that enters the action encoder alongside structured indicators. This is, to our knowledge, the first medical world model to formalise communication as a primary action.
    \item \textbf{A physiology-aware training objective.} We pair the LeWM next-latent-prediction and SIGReg backbone with three shape priors---slope consistency, local-continuity Huber, and a hinge-style large-jump penalty---that prevent representation collapse and clinically implausible step changes during multi-year rollout, and we train under a rollout-prefix protocol that matches the deployment regime.
    \item \textbf{A CKD eGFR case study demonstrating value.} We instantiate \cmwm{} for annual eGFR forecasting on a 2{,}232-patient nephrology cohort and show that the instantiated model outperforms a tuned GPT-5.5 structured-prompting baseline on dynamic-50\% rollout (test MAE 7.384 vs.\ 7.964; RMSE 10.256 vs.\ 11.069), with ablations confirming that the framework's communication-action channel drives the gain.
\end{itemize}

We emphasise that the architecture, loss design, and training protocol are domain-agnostic; the CKD instantiation supplies one concrete validation, and the same template applies to other chronic conditions whose management can be cast as periodic clinical state interleaved with structured and conversational interventions (e.g., glycated haemoglobin (HbA1c) in type 2 diabetes, blood pressure in hypertension, N-terminal pro--B-type natriuretic peptide (NT-proBNP) and ejection fraction in heart failure, forced expiratory volume in one second (FEV$_1$) in chronic obstructive pulmonary disease).

\section{Related Work}
\label{sec:related}

\subsection{World Models and Joint-Embedding Predictive Architectures}
World models compress an environment into a learned latent dynamics function that an agent can roll forward without further interaction \cite{ha2018worldmodels}. PlaNet \cite{hafner2019planet} introduced recurrent state-space models for pixel-based control; Dreamer-V3 \cite{hafner2023dreamerv3} generalised the paradigm across diverse domains using latent imagination. LeCun's energy-based view \cite{lecun2022pathway} reframed prediction as a joint-embedding problem, leading to image and video JEPAs \cite{assran2023ijepa}. LeWM \cite{maes2026leworldmodel} contributes a clean end-to-end JEPA training recipe by replacing collapse-prevention heuristics with SIGReg, a normality test on random one-dimensional projections of the latent distribution. \cmwm{} retains LeWM's next-latent-prediction and SIGReg backbone, but the observation space is periodic structured clinical state, the action space couples structured intervention indicators with semantic communication embeddings, and the supervision target is a physiological scalar rather than a control reward.

\subsection{EHR Representation Learning and Medical World Models}
RETAIN \cite{choi2016retain} demonstrated interpretable longitudinal modelling via reverse-time attention; BEHRT \cite{li2020behrt} and Med-BERT \cite{rasmy2021medbert} adapted transformer pretraining to structured electronic health records, and the public availability of MIMIC-IV \cite{johnson2023mimiciv} catalysed a generation of dynamic EHR models. A 2023 systematic review of deep models on EHR trajectories \cite{amirahmadi2023review} identified missingness, irregular sampling, external validation and interpretability as persistent obstacles. Most of these works remain discriminative. EHRWorld \cite{mu2026ehrworld} represents a recent pivot: it explicitly reframes EHR modelling as long-horizon causally-grounded patient simulation, and quantifies how LLM-only simulators drift after sequential interventions. \cmwm{} shares EHRWorld's emphasis on intervention-conditioned long-horizon consistency, but differs in two ways: it targets the periodic-observation chronic-disease regime rather than dense intensive-care-unit (ICU) style multivariate simulation, and it formalises free-text communication as a first-class action input rather than as context.

\subsection{Disease-Trajectory and Risk Modelling: a Reference Case Study}
Because we evaluate \cmwm{} on CKD, we briefly survey the CKD modelling landscape that frames our case study. Tangri's Kidney Failure Risk Equation (KFRE) \cite{tangri2011kfre} remains the reference clinical risk score; the Chronic Kidney Disease Epidemiology Collaboration (CKD-EPI) 2021 race-free creatinine equation \cite{inker2021ckdepi} is now standard for laboratory eGFR reporting. Recent trials---DAPA-CKD \cite{heerspink2020dapackd}, FIDELIO-DKD \cite{bakris2020fidelio}---and the Kidney Disease: Improving Global Outcomes (KDIGO) 2024 guideline \cite{kdigo2024guideline} reshaped pharmacological management, and methodological work has consolidated total eGFR slope as a surrogate endpoint for CKD progression \cite{inker2019gfrslope}. On the modelling side, time-dependent long short-term memory networks (LSTMs) \cite{yu2024tdlstm}, transformer time-to-event models \cite{ferro2024transformerckd}, and the recent KFDeep dynamic predictor \cite{ma2025kfdeep} have pushed survival-style prediction of kidney failure, while metabolic-flux digital twins \cite{barbieri2024digitaltwin} target diagnostic classification rather than trajectory. A separate strand of work predicts eGFR as a continuous value: deep-learning models from kidney ultrasound \cite{kuo2019kidneyus}, interpretable machine-learning models from routine laboratory features in CKD cohorts \cite{rojas2025egfrml}, and pretransplant-biopsy convolutional neural network (CNN) models that forecast posttransplant graft eGFR \cite{luo2021posttransplant} all report mean absolute error (MAE) as a primary evaluation metric. None of these models unrolls clinical state autoregressively under both medication and communication actions; this is the niche the CKD instantiation of \cmwm{} fills, and the gap exists analogously in the modelling literature of other chronic conditions.

\subsection{Irregular Time Series and Counterfactual Treatment Effects}
Periodic aggregation reduces but does not eliminate the irregular-sampling problem common to follow-up data. Latent ordinary differential equation (ODE) models \cite{rubanova2019latentode} and multi-time attention networks \cite{shukla2021mtand} provide continuous-time alternatives to discrete recurrence. For intervention-conditioned outcomes, the counterfactual recurrent network \cite{bica2020crn} and treatment-effect neural controlled differential equations (TE-CDE) \cite{seedat2022tecde} estimate treatment effects under time-dependent confounding. \cmwm{} is not a causal-inference method---it does not adjust for confounding---but it shares the goal of producing usable predictions under counterfactual action sequences, and its user-controllable action interface is designed so that downstream causal methods can be layered on top.

\section{The ChronoMedicalWorld Framework}
\label{sec:framework}

\subsection{Problem Formulation}
We consider a patient followed periodically. At each observation index $t=1,\ldots,T_i$, three vectors are recorded: a clinical state $x_{i,t}\in\mathbb{R}^{d_x}$ summarising the period (biomarkers, vitals, missingness flags, measurement counts), an action $a_{i,t}\in\mathbb{R}^{d_a}$ that may concatenate structured intervention indicators $a^m_{i,t}$ and a semantic embedding $a^c_{i,t}$ of any free-text exchange during the period, and a target-time covariate vector $\tau_{i,t}\in\mathbb{R}^{d_\tau}$ (e.g., age, gap since previous observation, follow-up volume). The patient also carries a static context $b_i$ such as sex and disease type.

Let
\begin{equation}
\mathcal{H}_{i,t}=\{(x_{i,1},a_{i,1},\tau_{i,1}),\ldots,(x_{i,t},a_{i,t},\tau_{i,t})\}.
\end{equation}
A \emph{medical world model} is a parametric map
\begin{equation}
\hat{y}_{i,t+k}=f_\theta\!\big(\mathcal{H}_{i,t},\,b_i,\,a_{i,t:t+k-1},\,\tau_{i,t+1:t+k}\big),\qquad k\ge 1,
\label{eq:problem}
\end{equation}
where $\hat{y}_{i,t+k}$ is a designated component (or transformation) of the clinical state---the \emph{target observable}---that we wish to forecast under candidate intervention sequences. At evaluation, $\hat{y}_{i,t+k}$ is written back into the corresponding slot of $x_{i,t+k}^{(1)}$ in the next step, so that no future ground-truth observable enters the rollout.

\subsection{Architecture}
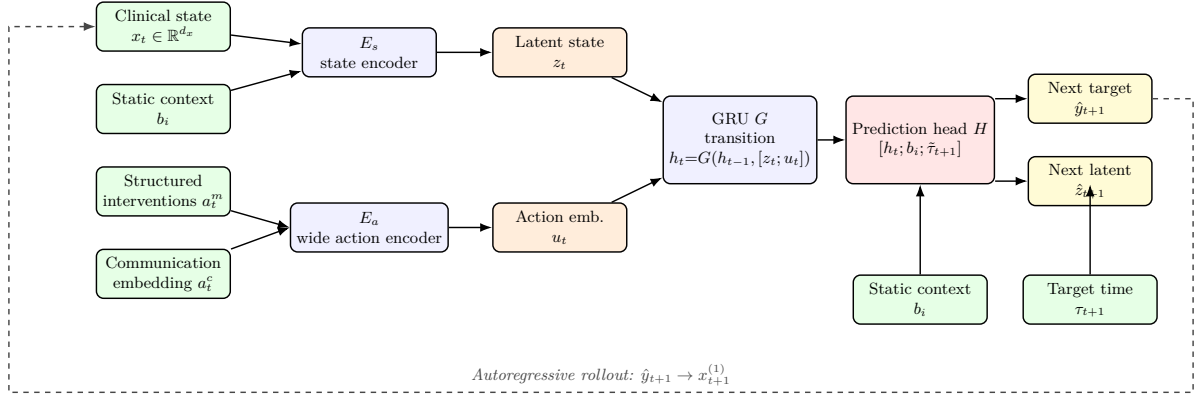
\begin{figure}[H]
\centering
\resizebox{0.95\textwidth}{!}{%
\begin{tikzpicture}[
    >=Latex,
    font=\small,
    every node/.style={align=center},
    cwbox/.style={draw, rounded corners=4pt, thick, minimum width=2.6cm, minimum height=0.95cm, fill=blue!6},
    cwinp/.style={draw, rounded corners=4pt, thick, minimum width=2.6cm, minimum height=0.95cm, fill=green!10},
    cwlat/.style={draw, rounded corners=4pt, thick, minimum width=2.6cm, minimum height=0.95cm, fill=orange!14},
    cwhead/.style={draw, rounded corners=4pt, thick, minimum width=2.7cm, minimum height=1.7cm, fill=red!10},
    cwout/.style={draw, rounded corners=4pt, thick, minimum width=2.4cm, minimum height=0.95cm, fill=yellow!20},
]

\node[cwinp] (x)  at (0, 3.2)  {Clinical state\\$x_t\in\mathbb{R}^{d_x}$};
\node[cwinp] (bL) at (0, 1.6)  {Static context\\$b_i$};
\node[cwinp] (am) at (0, 0.0)  {Structured\\interventions $a^m_t$};
\node[cwinp] (ac) at (0,-1.6)  {Communication\\embedding $a^c_t$};

\node[cwbox] (Es) at (4.0, 2.7) {$E_s$\\state encoder};
\node[cwbox] (Ea) at (4.0,-0.7) {$E_a$\\wide action encoder};

\node[cwlat] (z)  at (7.7, 2.7)  {Latent state\\$z_t$};
\node[cwlat] (u)  at (7.7,-0.7)  {Action emb.\\$u_t$};

\node[cwbox,minimum height=1.7cm,minimum width=2.7cm] (G) at (11.2, 1.0)
  {GRU $G$\\transition\\$h_t{=}G(h_{t-1},[z_t;u_t])$};

\node[cwhead] (H) at (14.7, 1.0) {Prediction head $H$\\$[h_t;b_i;\tilde\tau_{t+1}]$};

\node[cwout] (yhat) at (18.0, 1.8) {Next target\\$\hat y_{t+1}$};
\node[cwout] (zhat) at (18.0, 0.2) {Next latent\\$\hat z_{t+1}$};

\node[cwinp] (bR)   at (14.7,-2.1) {Static context\\$b_i$};
\node[cwinp] (tau)  at (18.0,-2.1) {Target time\\$\tau_{t+1}$};

\draw[->,thick] (x)  -- (Es);
\draw[->,thick] (bL) -- (Es);                       
\draw[->,thick] (Es) -- (z);
\draw[->,thick] (am) -- (Ea.west);
\draw[->,thick] (ac) -- (Ea.west);
\draw[->,thick] (Ea) -- (u);
\draw[->,thick] (z) -- (G);
\draw[->,thick] (u) -- (G);
\draw[->,thick] (G) -- (H);
\draw[->,thick] (H.east |- yhat) -- (yhat);
\draw[->,thick] (H.east |- zhat) -- (zhat);
\draw[->,thick] (bR)  -- (H.south -| bR);
\draw[->,thick] (tau) -- (H.south -| tau);

\coordinate (fb_NE) at (20.0, 1.8);
\coordinate (fb_SE) at (20.0,-3.9);
\coordinate (fb_SW) at (-3.0,-3.9);
\coordinate (fb_NW) at (-3.0, 3.2);

\draw[->,dashed,thick,gray!55!black]
  (yhat.east) -- (fb_NE) -- (fb_SE) -- (fb_SW) -- (fb_NW) -- (x.west);

\node[gray!55!black, font=\small\itshape, anchor=south]
  at (8.5,-3.9) {Autoregressive rollout: $\hat y_{t+1}\to x_{t+1}^{(1)}$};

\end{tikzpicture}%
}
\caption{The ChronoMedicalWorld Model (\cmwm) framework. State pathway: a periodic clinical state $x_t$ together with the static patient embedding $b_i$ is mapped by the state encoder $E_s$ into the latent state $z_t$. Action pathway: structured intervention indicators $a^m_t$ and a semantic embedding $a^c_t$ of within-period patient--health-coach communication are mapped jointly by the wide action encoder $E_a$ into the action embedding $u_t$. The recurrent transition module $G$ (implemented as a gated recurrent unit, GRU, in our case-study instantiation) propagates the fused stream $[z_t;u_t]$ into a hidden state $h_t$; the head $H$ consumes $[h_t;b_i;\tilde\tau_{t+1}]$ and produces both the next-period target observable $\hat y_{t+1}$ (e.g., eGFR, HbA1c, ejection fraction) and the next predicted latent $\hat z_{t+1}$. The static context $b_i$ is shown twice for layout clarity (left of $E_s$ and below $H$); it denotes the same input fed to both modules. The perimeter dashed loop indicates the autoregressive rollout used at evaluation, where $\hat y_{t+1}$ replaces the corresponding slot of the next-period clinical state.}
\label{fig:architecture}
\end{figure}

\cmwm{} instantiates Equation~\eqref{eq:problem} as a recurrent latent transition module with four components (Figure~\ref{fig:architecture}):
\begin{align}
b_i &= B(\mathrm{static}_i),\\
z_{i,t} &= E_s\!\big([\tilde{x}_{i,t};\,b_i]\big), \quad &&\text{(state encoder)} \\
u_{i,t} &= E_a\!\big([a^m_{i,t};\,a^c_{i,t}]\big), \quad &&\text{(wide action encoder)} \\
h_{i,t} &= G\!\big(h_{i,t-1},\,[z_{i,t};\,u_{i,t}]\big), \quad &&\text{(recurrent transition)} \\
(\hat{y}_{i,t+1},\,\hat{z}_{i,t+1}) &= H\!\big([h_{i,t};\,b_i;\,\tilde{\tau}_{i,t+1}]\big). \quad &&\text{(prediction head)}
\end{align}
Tildes denote standardisation using train-set statistics. The two output channels are deliberate: the explicit-prediction channel $\hat{y}_{i,t+1}$ supplies supervised, interpretable forecasts for the target observable, while the latent channel $\hat{z}_{i,t+1}$ enables JEPA-style next-latent training and downstream latent-space probing. The wide action encoder is a single feed-forward block over the full action vector; the framework also admits a split variant in which structured and text actions are routed through separate encoders, but in our experiments the wide variant proved more stable under multi-step rollout.

\subsection{Training Objective}
For minibatches drawn from rollout-prefix samples (Section~\ref{sec:rollout_prefix}), \cmwm{} minimises
\begin{equation}
\mathcal{L}=\mathcal{L}_{y}+\lambda_z\mathcal{L}_z+\lambda_{\mathrm{sig}}\mathcal{L}_{\mathrm{SIGReg}}+\lambda_\Delta\mathcal{L}_{\mathrm{slope}}+\lambda_c\mathcal{L}_{\mathrm{cont}}+\lambda_j\mathcal{L}_{\mathrm{jump}}.
\label{eq:total_loss}
\end{equation}
The supervised term penalises one-step error on the target observable in the standardised space:
\begin{equation}
\mathcal{L}_{y}=\frac{1}{|\mathcal{D}|}\sum_{(i,t)\in\mathcal{D}} \smooth\!\left(\hat{y}_{i,t+1}-y_{i,t+1}\right).
\end{equation}
The next-latent term aligns the predicted next latent with the encoded next observed state under stop-gradient on the target encoding, as in JEPA-style models \cite{maes2026leworldmodel,assran2023ijepa}:
\begin{equation}
\mathcal{L}_z=\frac{1}{|\mathcal{D}|}\sum_{(i,t)\in\mathcal{D}}\big\|\hat{z}_{i,t+1}-\sg\!\big(E_s\!\big([\tilde{x}_{i,t+1};b_i]\big)\big)\big\|_2^2.
\end{equation}

SIGReg \cite{maes2026leworldmodel} prevents trivial latent collapse by projecting a minibatch of latent codes onto $Q$ random unit directions and measuring the discrepancy between each one-dimensional empirical characteristic function $\hat\phi_q(t)=\hat C_q(t)+i\hat S_q(t)$ and the characteristic function $\phi^{\star}(t)=e^{-t^{2}/2}$ of a standard Gaussian:
\begin{align}
\hat C_q(t) &= \tfrac{1}{N}\textstyle\sum_{n=1}^{N}\cos\!\big(t\, q^{\top}z_n\big), \quad \hat S_q(t) = \tfrac{1}{N}\textstyle\sum_{n=1}^{N}\sin\!\big(t\, q^{\top}z_n\big), \\
\mathcal{L}_{\mathrm{SIGReg}} &= \frac{N}{Q}\sum_{q=1}^{Q}\sum_{k=0}^{K-1} w_k\,\Big[\big(\hat C_q(t_k)-\phi^{\star}(t_k)\big)^{2} + \hat S_q(t_k)^{2}\Big],
\label{eq:sigreg}
\end{align}
where $\{t_k\}_{k=0}^{K-1}$ is a uniform grid over $[0,T_{\max}]$, $w_k$ are Gaussian-windowed trapezoidal weights, $z_n$ is the $n$-th latent code in the minibatch, and $q$ is sampled from the uniform distribution on the unit sphere. We use $K=17$, $T_{\max}=3$, and $Q=64$. Equation~\eqref{eq:sigreg} is differentiable in $\theta$ and acts as an isotropy regulariser without exponential moving averages or auxiliary supervision.

To match the prior that chronic-disease physiology evolves smoothly and rarely with abrupt single-period jumps, we add three shape penalties. The slope-consistency term aligns the predicted one-period change with the observed one-period change:
\begin{equation}
\mathcal{L}_{\mathrm{slope}}=\frac{1}{|\mathcal{D}|}\sum_{(i,t)\in\mathcal{D}}\smooth\!\big(\,(\hat{y}_{i,t+1}-y_{i,t})-(y_{i,t+1}-y_{i,t})\,\big).
\end{equation}
Local continuity is enforced by a Huber loss between the predicted slope and zero (in standardised units), discouraging small but persistent biases that compound during rollout:
\begin{equation}
\mathcal{L}_{\mathrm{cont}}=\frac{1}{|\mathcal{D}|}\sum_{(i,t)\in\mathcal{D}}\mathrm{Huber}_{\delta_c}\!\big(\hat{y}_{i,t+1}-y_{i,t},\,0\big).
\end{equation}
Clinically implausible single-period jumps are penalised by a one-sided hinge:
\begin{equation}
\mathcal{L}_{\mathrm{jump}}=\frac{1}{|\mathcal{D}|}\sum_{(i,t)\in\mathcal{D}}\max\!\big(0,\;|\hat{y}_{i,t+1}-y_{i,t}|-\delta_j\big)^{2}.
\end{equation}
The three priors have complementary roles: $\mathcal{L}_{\mathrm{slope}}$ targets the \emph{direction} of progression, $\mathcal{L}_{\mathrm{cont}}$ smooths short-term oscillation in any direction, and $\mathcal{L}_{\mathrm{jump}}$ is silent inside a clinically acceptable band $[-\delta_j,\delta_j]$ but penalises only the part of the step beyond $\delta_j$. The thresholds $\delta_c,\delta_j$ are framework hyperparameters whose calibration depends on the units and scale of the target observable.

\subsection{Rollout-Prefix Training Protocol}
\label{sec:rollout_prefix}
A long-standing source of multi-step error in autoregressive sequence models is the train--evaluation mismatch between teacher-forced next-step training and free-running rollout deployment. \cmwm{} addresses this by replacing standard one-step training with a \emph{rollout-prefix} regime. For a patient with $T_i$ observations, every prefix length $c\in[c_{\min}, T_i)$ generates one training trajectory in which the model unrolls forward for up to $H$ steps; predicted target observables are fed back into the next step's clinical state in place of the corresponding ground-truth slot, exactly as at evaluation; and per-step losses are aggregated with a horizon-decay factor so that nearer-horizon steps dominate optimisation while later-horizon steps still contribute. This protocol makes the model robust to its own prediction error and is, in our experiments, more important for closed-loop performance than any single architectural choice.

\subsection{Inference-Time Stabilisation}
At inference, \cmwm{} supports an optional first-step anchor: the first prediction is convex-combined with a recent-trend extrapolation from the patient's own history, and the resulting first-step change is clipped to a clinically plausible magnitude. This is a deployment-time stabiliser independent of training; it eliminates an initial jump that occasionally appears at the boundary between observed history and pure-rollout forecasting and that compounds over later horizons.

\section{Case Study: Annual eGFR Trajectory Forecasting in CKD}
\label{sec:casestudy}
To substantiate the value of \cmwm{}, we instantiate the framework on a non-trivial chronic-disease setting---annual eGFR trajectory forecasting in CKD---and compare it against a contemporary large language model under matched evaluation protocol. The case study is intended as evidence of the framework's value; the framework itself remains domain-agnostic.

\subsection{Cohort}
The study used a private KidneyOnline platform follow-up cohort. Data sources included serum-creatinine--derived eGFR computed by the CKD-EPI 2021 race-free equation \cite{inker2021ckdepi}; demographics and pathology; medication exposures; 24-hour urine protein; patient-reported home-measured blood pressure; and patient--health-coach chat transcripts collected through the platform messaging interface. Records were aggregated by calendar year to produce annual feature vectors. Inclusion required at least five annual observations and a first-year eGFR above 30 mL/min/1.73\,m\textsuperscript{2}. Patients were split at the patient level. The final cohort comprised 2{,}232 patients and 15{,}070 patient-years.

\begin{table}[htbp]
\centering
\caption{KidneyOnline cohort characteristics, feature dimensions, and missingness for the CKD case study. Continuous variables are reported as mean (standard deviation, SD) or median [interquartile range, IQR]. MAP, mean arterial pressure.}
\label{tab:dataset_summary}
\small
\setlength{\tabcolsep}{4pt}
\begin{tabularx}{\textwidth}{p{0.29\textwidth}p{0.24\textwidth}Y}
\toprule
Category & Statistic & Value \\
\midrule
Cohort size & Patients / patient-years & 2{,}232 / 15{,}070 \\
Follow-up & Calendar years; years per patient & 1999--2026; 6.0 [5.0--8.0] \\
Baseline demographics & Age; sex & 35.2 (10.0) years; 1{,}139 female (51.0\%), 1{,}093 male (49.0\%) \\
Baseline renal status & eGFR; 24-h urine protein; MAP & 94.7 (26.5); 0.5 [0.3--1.1] g/day; 85.9 (8.5) mmHg \\
Pathology & Most common diagnoses & IgA nephropathy 46.4\%, unbiopsied/unspecified 19.4\%, membranous nephropathy 7.2\% \\
Static features & Variables / pathology categories & 2 variables; 50 pathology categories \\
Dynamic clinical state & Annual state dimensions & 9 dimensions \\
Target-time covariates & Dimensions & 6 dimensions \\
Structured intervention actions & Dimensions & 62 binary features (12 drug classes + 50 frequent agents) \\
Communication actions & Dimensions and coverage & 256-d annual text embedding; 10{,}045 patient-years with communication \\
Communication volume & Messages and characters & 805{,}327 messages and 46.4 M characters; nonzero years had 37 [8--98] messages and 2{,}442 [603--5{,}805] characters \\
Missingness & Patient-year missing rate & eGFR 0\% by construction; urine protein 28.2\%; blood pressure 40.0\%; no communication 33.3\% \\
\bottomrule
\end{tabularx}
\end{table}

\begin{table}[htbp]
\centering
\caption{Patient-level data split and evaluation sample counts for the CKD case study.}
\label{tab:data_split}
\small
\setlength{\tabcolsep}{4pt}
\resizebox{\textwidth}{!}{%
\begin{tabular}{lrrrr}
\toprule
Split & Patients & Next-year samples & Dynamic rollout points & Dynamic rollout patients \\
\midrule
Train & 1{,}562 & 5{,}306 & 4{,}875 & 1{,}562 \\
Validation & 334 & 1{,}110 & 1{,}020 & 334 \\
Test & 336 & 1{,}177 & 1{,}078 & 336 \\
\bottomrule
\end{tabular}%
}
\end{table}

\subsection{Instantiation: State, Action, Time, and Target}
The target observable is annual mean eGFR. The clinical state vector $x_t\in\mathbb{R}^{9}$ collects mean eGFR, mean age, log-transformed 24-hour urine total protein, median mean arterial pressure (MAP), measurement counts for eGFR / urine protein / blood pressure, and missingness flags for urine protein and blood pressure. The target-time vector $\tau_t\in\mathbb{R}^{6}$ contains age, gap to the previous annual visit, years since baseline, and the same measurement counts. Structured intervention actions $a^m_t\in\{0,1\}^{62}$ cover 12 drug classes (steroid, calcineurin inhibitor, mycophenolate mofetil (MMF), other immunosuppressant, renin--angiotensin system (RAS) blocker, sodium--glucose cotransporter-2 (SGLT2) inhibitor, mineralocorticoid receptor antagonist (MRA), uric-acid lowering agent, statin, calcium channel blocker (CCB), $\beta$-blocker, broad CKD support) and 50 frequent individual agents, encoded as year-level exposure indicators. The communication action $a^c_t\in\mathbb{R}^{256}$ is obtained by sorting all patient--health-coach messages of a patient-year chronologically, concatenating them into one annual transcript, and embedding with OpenAI's \texttt{text-embedding-3-small} model truncated to 256 dimensions \cite{openai2024embeddings}; long transcripts are embedded in overlapping chunks and averaged with character-count weights; years without communication receive a zero communication vector. The full action vector $a_t\in\mathbb{R}^{318}$ concatenates the structured and communication components.

\subsection{Training Configuration}
We train \cmwm{} with AdamW under the rollout-prefix protocol of Section~\ref{sec:rollout_prefix} and select on validation dynamic-50\% rollout MAE. Table~\ref{tab:training} summarises the configuration used for the case study.

\begin{table}[htbp]
\centering
\caption{Training configuration of the best \cmwm{} variant on the CKD case study (wide action encoder, communication embedding).}
\label{tab:training}
\small
\setlength{\tabcolsep}{4pt}
\begin{tabularx}{\textwidth}{p{0.40\textwidth}Y}
\toprule
Item & Value \\
\midrule
State / action / target-time dim. & 9 / 318 / 6 \\
Architecture & state encoder, wide action encoder, GRU dynamics \\
Hidden / latent / static / action emb. & 256 / 128 / 64 / 128 \\
Total trainable parameters & 790{,}081 ($\approx$0.79\,M; GRU 0.395\,M, prediction heads 0.217\,M, encoders 0.176\,M) \\
Context length / minimum history / max horizon & 6 / 3 / 8 \\
Optimiser & AdamW, lr $6\times10^{-4}$, weight decay $8\times10^{-5}$ \\
Dropout & 0.05 \\
Loss weights in Eq.~\eqref{eq:total_loss} & $\lambda_z{=}0.08,\ \lambda_{\mathrm{sig}}{=}0.008,\ \lambda_\Delta{=}0.2,\ \lambda_c{=}0.3,\ \lambda_j{=}0.2$ \\
Continuity / jump thresholds & $\delta_c{=}0.5,\ \delta_j{=}10$ eGFR units \\
SIGReg grid / projections & $K{=}17$ knots over $[0,3]$, $Q{=}64$ \\
Selection metric / best epoch & validation dynamic-50\% rollout MAE / 5 \\
Rollout-time stabilisation & first-step anchor weight 1.0, anchor jump cap $\pm5$ eGFR units \\
\bottomrule
\end{tabularx}
\end{table}

\subsection{GPT-5.5 Baseline}
We probe GPT-5.5 through structured prompting on the same history, future non-eGFR covariates and intervention descriptions used by \cmwm{}. The prompt restricts the model to outputting only future annual eGFR values and explicitly withholds future ground-truth eGFR. For each validation and test patient, the context length matches the dynamic-50\% rollout protocol ($\max(3,\lfloor 0.5\,T_i\rfloor)$). Predictions are de-duplicated per (patient, target-year) before scoring.

\section{Results}

\subsection{\cmwm{} vs.\ GPT-5.5 on the CKD Case Study}
We report mean absolute error (MAE) and root-mean-square error (RMSE) on the predicted vs.\ observed annual eGFR. MAE is the de facto standard summary metric for continuous-eGFR prediction in prior CKD modelling work---used by imaging-based deep models \cite{kuo2019kidneyus}, interpretable machine-learning models on routine laboratory features \cite{rojas2025egfrml}, and pretransplant-biopsy CNN models for posttransplant graft eGFR \cite{luo2021posttransplant}---so it allows our numbers to be read against the broader literature without re-scaling. Table~\ref{tab:gpt_comparison} reports dynamic-50\% history rollout metrics on the held-out validation and test splits, plus the sample-weighted overall summary. \cmwm{} outperforms GPT-5.5 across all rows. The relative improvements are largest in RMSE, indicating that the explicit medical world model is particularly effective at suppressing the long-tail late-horizon errors an LLM-only simulator accumulates after several rollout steps---consistent with the behaviour reported by EHRWorld \cite{mu2026ehrworld}.

\begin{table}[htbp]
\centering
\caption{Dynamic-50\% history rollout comparison between \cmwm{} (CKD instantiation) and GPT-5.5 on the CKD case study. ``Overall'' is the sample-weighted validation and test summary.}
\label{tab:gpt_comparison}
\small
\setlength{\tabcolsep}{4pt}
\resizebox{\textwidth}{!}{%
\begin{tabular}{lrrrrrrr}
\toprule
Split & $n$ & \cmwm{} MAE & \cmwm{} RMSE & GPT-5.5 MAE & GPT-5.5 RMSE & $\Delta$ MAE & $\Delta$ RMSE \\
\midrule
Validation & 1{,}020 & 7.310 & 9.851  & 7.997 & 11.128 & $-$8.59\% & $-$11.48\% \\
Test       & 1{,}078 & 7.384 & 10.256 & 7.964 & 11.069 & $-$7.28\% & $-$7.35\% \\
Overall    & 2{,}098 & 7.348 & 10.061 & 7.980 & 11.098 & $-$7.92\% & $-$9.34\% \\
\bottomrule
\end{tabular}%
}
\end{table}

\subsection{Action-Encoder Ablations}
Table~\ref{tab:ablation} reports four communication-aware variants under the same dynamic-50\% protocol. The wide encoder ingesting the full annual communication embedding is the best long-horizon model. A split structured/communication encoder slightly improves next-year MAE but degrades rollout, suggesting that stronger modular separation increases compounding variance in closed-loop forecasts. Replacing the semantic embedding by raw communication-intensity statistics (message count, character count, chunk count) does not recover the gain---a strong indication that the predictive signal lies in semantic content rather than volume, supporting the design choice to treat communication as a semantic action rather than a volumetric covariate.

\begin{table}[htbp]
\centering
\caption{Ablations of the action encoder on the CKD case study. ``Test MAE/RMSE'' are next-year metrics; ``Test rollout MAE'' uses the dynamic-50\% protocol; ``Fixed3 MAE'' anchors on exactly the first three years.}
\label{tab:ablation}
\small
\setlength{\tabcolsep}{4pt}
\resizebox{\textwidth}{!}{%
\begin{tabular}{lrrrrr}
\toprule
Model variant & Action dim. & Test MAE & Test RMSE & Test rollout MAE & Fixed3 MAE \\
\midrule
Communication embedding with wide encoder & 318 & 11.046 & 15.860 & \textbf{7.384} & \textbf{7.697} \\
Split structured / communication encoder & 321 & \textbf{10.953} & \textbf{15.834} & 7.526 & 7.768 \\
Communication intensity with wide encoder & 321 & 11.142 & 15.853 & 7.612 & 7.946 \\
Compact split encoder & 321 & 11.313 & 16.149 & 7.796 & 8.129 \\
\bottomrule
\end{tabular}%
}
\end{table}

A finer message-type ablation (dialogue-only vs.\ system-only communication embeddings, identical training otherwise) showed that dialogue messages reproduce essentially all of the gain: dynamic rollout MAE is reduced by 4.78\% relative to the no-communication baseline (compared with 4.95\% for the full-communication embedding), whereas system messages alone reduce MAE by only 2.04\%. The predictive value of the communication-action channel is therefore dominated by genuine patient--health-coach conversation, not by platform notifications.

\section{Discussion}

\subsection{Why an Explicit Medical World Model Outperforms an LLM Simulator}
GPT-5.5 has broad medical knowledge and reasons coherently over a single prompt, but is not explicitly trained to maintain a persistent latent patient state under repeated intervention-conditioned updates. In multi-year rollout this manifests as over-linearised decline or drift after an early misstep, mirroring the failure mode that EHRWorld \cite{mu2026ehrworld} reports more broadly for LLM-only clinical simulators. \cmwm{} carries an explicit recurrent hidden state and is optimised in the same closed-loop regime that defines the evaluation: every training trajectory feeds the model's own prediction back as input. The case-study performance gap is therefore larger on multi-step rollout than on next-year prediction, exactly the empirical signature we expect when long-horizon state consistency is the bottleneck. It is also worth noting that this is achieved at a fraction of the parameter budget: the CKD instance of \cmwm{} has roughly 0.79\,M trainable parameters (Table~\ref{tab:training}), several orders of magnitude smaller than the GPT-5.5 baseline, which is consistent with the broader observation that specialised, intervention-conditioned dynamics models can outperform much larger generalist simulators in long-horizon clinical tasks.

\subsection{Value of the Framework to Medical World Modelling}
\cmwm{} contributes four elements to the emerging medical-world-model literature.
\emph{First}, it specialises the JEPA next-latent-prediction and SIGReg recipe from raw-pixel control \cite{maes2026leworldmodel} to structured longitudinal clinical data, showing that the isotropy regulariser is useful beyond vision.
\emph{Second}, it formalises within-period patient--health-coach text exchange as a first-class action channel, extending the medical-world-model action interface beyond drug-and-procedure indicators. Because adherence cues, lifestyle coaching, and symptom reports surface primarily in conversation rather than structured fields, this design is well aligned with how chronic-disease care is actually delivered.
\emph{Third}, the rollout-prefix training protocol provides a concrete recipe for closing the train--evaluation gap in clinical autoregressive models, a generic source of long-horizon drift.
\emph{Fourth}, the user-controllable action interface, combined with first-step anchoring, makes the framework directly usable in a counterfactual ``what-if'' interface---a foundation for personalised follow-up planning, intervention-aware decision support, and clinical digital-twin systems \cite{barbieri2024digitaltwin}.

\subsection{Generalisation Beyond CKD}
None of \cmwm{}'s components is specific to nephrology. The state vector $x_t$ accommodates any structured periodic biomarker panel; the action vector $a_t$ accommodates any combination of structured intervention indicators and free-text-derived semantic embeddings; the target observable $\hat y_{t+1}$ can be any continuous physiological scalar (or, with a different head, any discrete event or vector-valued state). The same template therefore applies, with appropriate calibration of $\delta_c$ and $\delta_j$ to the target's units, to long-horizon trajectory modelling of, for example, HbA1c in type 2 diabetes, blood pressure in hypertension, left-ventricular ejection fraction and NT-proBNP in heart failure, FEV$_1$ in chronic obstructive pulmonary disease, and disease activity scores in autoimmune diseases. The CKD case study should be read as a feasibility and value demonstration on a setting where (i) we have a longitudinal cohort with both structured therapy and rich patient--health-coach conversation, and (ii) a contemporary LLM baseline is available for direct comparison; analogous instantiations in other chronic conditions are an immediate follow-up.

\subsection{Limitations}
The case-study cohort is single-source, single-platform, and skews young (median age 35); external validation in older and multi-centre populations is required before clinical deployment, and a similar validation regime will be needed for each future disease instantiation. Structured intervention actions are annual binary indicators that do not encode dose, duration, adherence, or confounding by indication; \cmwm{} is not a causal estimator, and comparison with methods that explicitly adjust for time-dependent confounding \cite{bica2020crn,seedat2022tecde} is left to future work. The communication embedding is a period-level summary and does not separate health-coach advice, patient symptom reports, and automated system notifications inside a single window. Finally, the GPT-5.5 baseline is a structured-prompting baseline rather than a domain-fine-tuned LLM; the gap against a clinically-fine-tuned LLM is an open question.

\section{Conclusion}
We proposed \cmwm{} (ChronoMedicalWorld Model), a general action-conditioned latent world-model framework for learning patient trajectories from longitudinal care data. \cmwm{} couples a clinical-state encoder, a wide action encoder admitting both structured intervention indicators and free-text communication embeddings, and a recurrent latent transition module, and is trained under a six-term physiology-aware objective and a closed-loop rollout-prefix protocol. As a concrete case study, we instantiated \cmwm{} for annual eGFR trajectory forecasting in chronic kidney disease and demonstrated that the CKD instance outperforms a tuned GPT-5.5 structured-prompting baseline on dynamic-50\% rollout (test MAE 7.384 vs.\ 7.964; RMSE 10.256 vs.\ 11.069), with the gain dominated by the dialogue portion of patient--health-coach communication. The framework is not specific to CKD; the architecture, loss, and training protocol apply to any chronic-disease management process that can be cast as periodic clinical state interleaved with structured and conversational interventions, and \cmwm{} is offered as a medical-world-model template for that broader class of problems.

\appendix

\section{Case Studies}
\begin{figure}[H]
\centering
\begin{subfigure}{0.48\textwidth}
\centering
\begin{tikzpicture}
\begin{axis}[
width=\linewidth,height=4.9cm,
xlabel=Year,ylabel=eGFR,grid=major,
title={Case A},
ymin=13,ymax=65,
legend to name=casestudylegend,
legend columns=4,
legend style={font=\scriptsize, draw=none, /tikz/every even column/.append style={column sep=0.35cm}},tick label style={font=\scriptsize}, label style={font=\scriptsize}, title style={font=\small}
]
\addplot+[black, thick, mark=*, mark options={fill=black}] coordinates {(2014,54.1) (2015,56.8) (2016,54.5) (2017,55.5) (2018,49.0) (2019,50.3)};
\addplot+[green!45!black, thick, mark=diamond*, mark options={fill=green!45!black}] coordinates {(2020,43.0) (2021,32.1) (2022,33.2) (2023,39.1) (2024,23.5) (2025,20.9)};
\addplot+[gray!70, densely dashed, thick, mark=none, forget plot] coordinates {(2019,50.3) (2020,43.0)};
\addplot+[gray!70, densely dashed, thick, mark=none, forget plot] coordinates {(2019,50.3) (2020,47.7)};
\addplot+[gray!70, densely dashed, thick, mark=none, forget plot] coordinates {(2019,50.3) (2020,49.6)};
\addplot+[blue!70!black, thick, mark=square*] coordinates {(2020,47.7) (2021,40.0) (2022,38.2) (2023,33.7) (2024,31.4) (2025,31.5)};
\addplot+[red!75!black, thick, mark=triangle*] coordinates {(2020,49.6) (2021,49.2) (2022,47.8) (2023,46.9) (2024,46.1) (2025,45.0)};
\legend{history,actual future,CMWM,GPT-5.5}
\end{axis}
\end{tikzpicture}
\end{subfigure}
\hfill
\begin{subfigure}{0.48\textwidth}
\centering
\begin{tikzpicture}
\begin{axis}[
width=\linewidth,height=4.9cm,
xlabel=Year,ylabel=eGFR,grid=major,
title={Case B},
ymin=41,ymax=93,
tick label style={font=\scriptsize}, label style={font=\scriptsize}, title style={font=\small}
]
\addplot+[black, thick, mark=*, mark options={fill=black}] coordinates {(2017,74.5) (2018,78.5) (2019,84.9) (2020,67.0)};
\addplot+[green!45!black, thick, mark=diamond*, mark options={fill=green!45!black}] coordinates {(2021,70.9) (2022,61.6) (2023,62.5) (2024,53.5) (2025,52.4)};
\addplot+[gray!70, densely dashed, thick, mark=none, forget plot] coordinates {(2020,67.0) (2021,70.9)};
\addplot+[gray!70, densely dashed, thick, mark=none, forget plot] coordinates {(2020,67.0) (2021,62.0)};
\addplot+[gray!70, densely dashed, thick, mark=none, forget plot] coordinates {(2020,67.0) (2021,68.2)};
\addplot+[blue!70!black, thick, mark=square*] coordinates {(2021,62.0) (2022,61.9) (2023,56.4) (2024,51.5) (2025,48.7)};
\addplot+[red!75!black, thick, mark=triangle*] coordinates {(2021,68.2) (2022,66.7) (2023,64.8) (2024,64.9) (2025,65.6)};
\end{axis}
\end{tikzpicture}
\end{subfigure}

\par\medskip

\begin{subfigure}{0.48\textwidth}
\centering
\begin{tikzpicture}
\begin{axis}[
width=\linewidth,height=4.9cm,
xlabel=Year,ylabel=eGFR,grid=major,
title={Case C},
ymin=24,ymax=99,
tick label style={font=\scriptsize}, label style={font=\scriptsize}, title style={font=\small}
]
\addplot+[black, thick, mark=*, mark options={fill=black}] coordinates {(2018,90.8) (2019,76.5) (2020,67.6) (2021,61.5)};
\addplot+[green!45!black, thick, mark=diamond*, mark options={fill=green!45!black}] coordinates {(2022,58.0) (2023,57.4) (2024,59.9) (2025,58.3) (2026,54.5)};
\addplot+[gray!70, densely dashed, thick, mark=none, forget plot] coordinates {(2021,61.5) (2022,58.0)};
\addplot+[gray!70, densely dashed, thick, mark=none, forget plot] coordinates {(2021,61.5) (2022,56.5)};
\addplot+[gray!70, densely dashed, thick, mark=none, forget plot] coordinates {(2021,61.5) (2022,55.8)};
\addplot+[blue!70!black, thick, mark=square*] coordinates {(2022,56.5) (2023,60.0) (2024,55.8) (2025,52.8) (2026,50.8)};
\addplot+[red!75!black, thick, mark=triangle*] coordinates {(2022,55.8) (2023,49.9) (2024,43.8) (2025,38.6) (2026,31.7)};
\end{axis}
\end{tikzpicture}
\end{subfigure}
\hfill
\begin{subfigure}{0.48\textwidth}
\centering
\begin{tikzpicture}
\begin{axis}[
width=\linewidth,height=4.9cm,
xlabel=Year,ylabel=eGFR,grid=major,
title={Case D},
ymin=8,ymax=67,
tick label style={font=\scriptsize}, label style={font=\scriptsize}, title style={font=\small}
]
\addplot+[black, thick, mark=*, mark options={fill=black}] coordinates {(2020,59.5) (2021,52.1) (2022,44.0)};
\addplot+[green!45!black, thick, mark=diamond*, mark options={fill=green!45!black}] coordinates {(2023,42.7) (2024,40.4) (2025,36.9) (2026,31.0)};
\addplot+[gray!70, densely dashed, thick, mark=none, forget plot] coordinates {(2022,44.0) (2023,42.7)};
\addplot+[gray!70, densely dashed, thick, mark=none, forget plot] coordinates {(2022,44.0) (2023,39.0)};
\addplot+[gray!70, densely dashed, thick, mark=none, forget plot] coordinates {(2022,44.0) (2023,36.4)};
\addplot+[blue!70!black, thick, mark=square*] coordinates {(2023,39.0) (2024,42.1) (2025,34.9) (2026,35.0)};
\addplot+[red!75!black, thick, mark=triangle*] coordinates {(2023,36.4) (2024,29.1) (2025,22.3) (2026,16.1)};
\end{axis}
\end{tikzpicture}
\end{subfigure}

\par\medskip\ref{casestudylegend}
\caption{Representative dynamic-50\% test rollouts on the CKD case study in which \cmwm{} stays closer to the held-out future eGFR trajectory than GPT-5.5.}
\label{fig:case_study}
\end{figure}
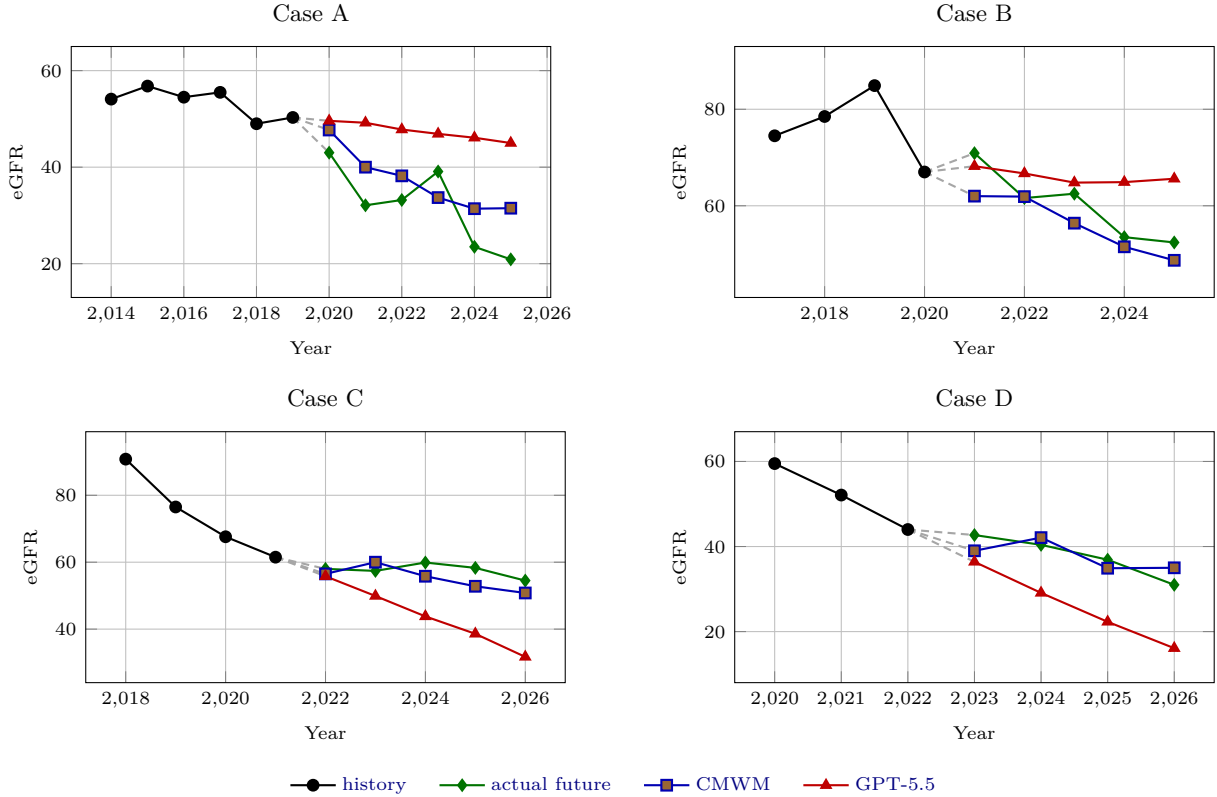

Figure~\ref{fig:case_study} shows four de-identified test cases from the CKD instantiation in which \cmwm{} outperformed GPT-5.5. The black curve is the historical context used for prediction; the green curve is the held-out future trajectory; coloured forecast curves start from the prediction horizon. Light dashed connectors link the last observed historical point to the first held-out future point and to the first prediction from each model. Per-case mean absolute errors are reported in Table~\ref{tab:case_errors}.

\begin{table}[H]
\centering
\caption{Mean absolute error of the four case-study patients in Figure~\ref{fig:case_study}.}
\label{tab:case_errors}
\begin{tabular}{lrrr}
\toprule
Case & Forecast years & \cmwm{} MAE & GPT-5.5 MAE \\
\midrule
A & 6 & 6.915 & 15.464 \\
B & 5 & 4.192 & 6.929 \\
C & 5 & 3.480 & 13.663 \\
D & 4 & 2.877 & 11.783 \\
\bottomrule
\end{tabular}
\end{table}

\clearpage
\bibliographystyle{unsrt}
\bibliography{egfr_world_model_refs}
\end{document}